%% file: acl_latex.tex
\documentclass[11pt]{article}
\usepackage[utf8]{inputenc}

\usepackage[final]{acl}

\usepackage{times}
\usepackage{latexsym}

\usepackage[T1]{fontenc}
\usepackage{enumitem}

\usepackage{microtype}

\usepackage{inconsolata}

\usepackage{graphicx}
\usepackage{booktabs}      
\usepackage{multirow}     
\usepackage{arydshln}      
\usepackage{amsmath}
\usepackage[table]{xcolor}

\usepackage{float}
\usepackage{placeins}
\usepackage{amssymb}

\newcommand{\ourmodel}{LRAS}
\usepackage{listings}
\usepackage{xcolor}
\usepackage{tcolorbox}
\tcbuselibrary{listings, breakable, skins}

\definecolor{promptbg}{RGB}{248, 249, 252}
\definecolor{promptframe}{RGB}{180, 185, 200}
\definecolor{titlebg}{RGB}{70, 90, 120}
\definecolor{keywordcolor}{RGB}{30, 100, 170}
\definecolor{stringcolor}{RGB}{40, 120, 80}
\definecolor{commentcolor}{RGB}{130, 130, 140}

\newtcblisting{promptbox}[2][]{
    enhanced,
    colback=promptbg,
    colframe=promptframe,
    colbacktitle=titlebg,
    coltitle=white,
    fonttitle=\bfseries\small,
    title={#2},
    boxrule=0.6pt,
    arc=3pt,
    left=8pt,
    right=8pt,
    top=4pt,
    bottom=4pt,
    toptitle=3pt,
    bottomtitle=3pt,
    listing only,
    listing options={
        basicstyle=\ttfamily\scriptsize\linespread{1.1}\selectfont,
        breaklines=true,
        breakatwhitespace=false,
        columns=fullflexible,
        keepspaces=true,
        showstringspaces=false,
        commentstyle=\color{commentcolor}\itshape,
        keywordstyle=\color{keywordcolor},
        stringstyle=\color{stringcolor},
        inputencoding=utf8,
        extendedchars=true,
        xleftmargin=0pt,
        xrightmargin=0pt,
    },
    #1
}

\usepackage{tcolorbox}
\tcbuselibrary{breakable, skins, listings}

\newtcolorbox{casestudybox}[2][]{
    enhanced,
    width=\textwidth,  
    colback=promptbg,
    colframe=promptframe,
    colbacktitle=titlebg,
    coltitle=white,
    fonttitle=\bfseries\small,
    title={#2},
    boxrule=0.6pt,
    arc=3pt,
    left=8pt,
    right=8pt,
    top=6pt,
    bottom=6pt,
    toptitle=3pt,
    bottomtitle=3pt,
    #1
}

\title{LRAS: Advanced Legal Reasoning with Agentic Search}

\author{
\textbf{Yujin Zhou}$^{1*}$, \textbf{Chuxue Cao}$^{1,2*}$, \textbf{Jinluan Yang}$^{*}$ \\
\textbf{Lijun Wu}$^2$, \textbf{Conghui He}$^2$, \textbf{Sirui Han}$^{1\dag}$, \textbf{Yike Guo}$^{1\dag}$\\
$^1$Hong Kong University of Science and Technology \\
$^2$Shanghai Artificial Intelligence Laboratory \\
\texttt{yzhouha@connect.ust.hk}
}

\begin{document}
\maketitle
{
\renewcommand{\thefootnote}{\fnsymbol{footnote}}
\footnotetext[1]{Equal contribution. $^\dag$Corresponding author.}
}


\input{sections/abstract}

\section{Introduction}\label{sec:introduction}
\input{sections/introduction}

\section{Related Works}\label{sec:related_works}
\input{sections/related_work}

\section{Preliminary and Motivation}\label{sec:pre_and_motivation}
\input{sections/pre_and_moti}

\section{Methodology}\label{sec:method}
\input{sections/approach}

\section{Experiments}\label{sec:exp}

\input{sections/experiments}

\section{Conclusion}\label{sec:con}
\input{sections/conclusion}

\section{Limitations}

\input{sections/limitations}

\section{Ethical considerations}
The dataset construction relied exclusively on publicly available or appropriately licensed resources, ensuring full compliance with licensing requirements. No personally identifiable or sensitive data were collected during the study. Human annotators were only involved in the verification of the dataset quality, and all annotators were recruited from a pool of qualified experts with backgrounds in law or computer science.

\bibliography{custom}

\newpage

\appendix

\input{sections/appendix}

\end{document}

%% file: sections/abstract.tex
\begin{abstract}
While Large Reasoning Models (LRMs) have demonstrated exceptional logical capabilities in mathematical domains, their application to the legal field remains hindered by the strict requirements for procedural rigor and adherence to legal logic. Existing legal LLMs, which rely on "closed-loop reasoning" derived solely from internal parametric knowledge, frequently suffer from lack of self-awareness regarding their knowledge boundaries, leading to confident yet incorrect conclusions. To address this challenge, we present Legal Reasoning with Agentic Search (LRAS), the first framework designed to transition legal LLMs from static and parametric "closed-loop thinking" to dynamic and interactive "Active Inquiry". By integrating Introspective Imitation Learning  and Difficulty-aware Reinforcement Learning, LRAS enables LRMs to identify knowledge boundaries and handle legal reasoning complexity. Empirical results demonstrate that LRAS outperforms state-of-the-art baselines by 8.2-32\%, with the most substantial gains observed in tasks requiring deep reasoning with reliable knowledge. We will release our data and models for further exploration soon.

\end{abstract}

%% file: sections/introduction.tex
\begin{figure}[t]
    \centering
    \includegraphics[width=\columnwidth]{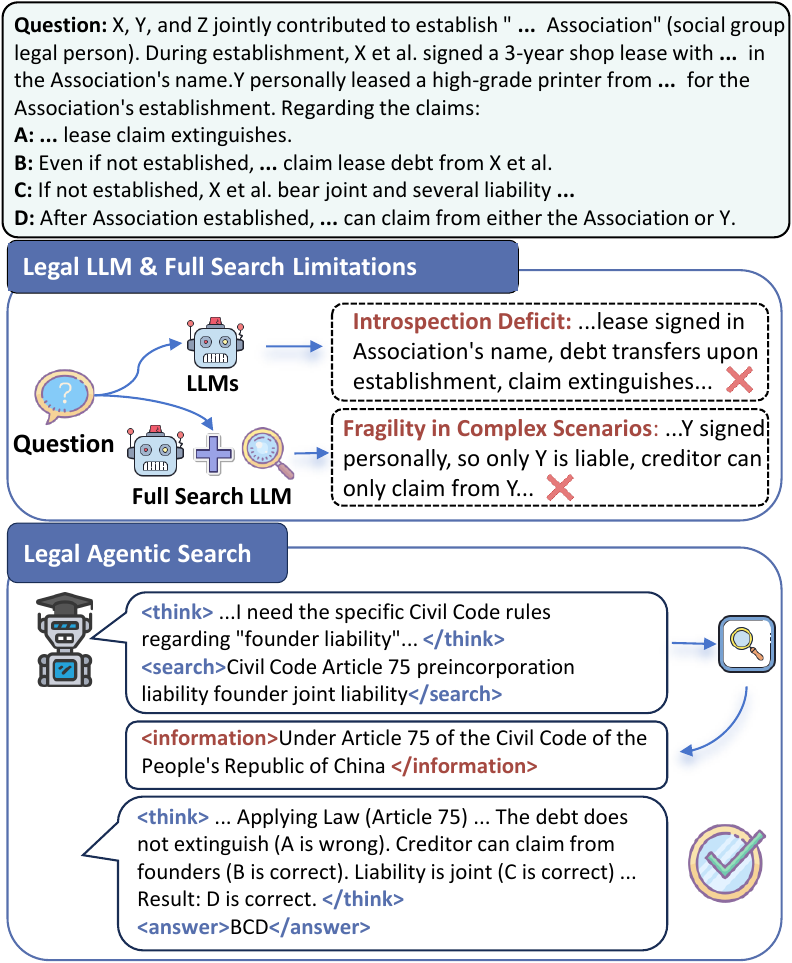}
    \caption{Comparison of legal LLMs, traditional search, and our proposed legal agentic search framework. Conventional legal LLMs suffer from limited introspection capabilities, failing to recognize their knowledge boundaries. In contrast, LRAS systematically identifies knowledge gaps, retrieves relevant statutes with precision, and generates well-grounded legal conclusions.}
    \label{fig:legal_comparison}
\vspace{-4mm}
\end{figure}

Large Reasoning Models (LRMs) have achieved great success in expanding the boundaries of complex logical problem-solving in mathematical and symbolic domains~\citep{yang2025qwen3, guo2025deepseek, cao2025towards,chen2025can}.  However, despite these advancements, existing approaches fall short in legal scenarios, which demand extreme precision and strict adherence to procedural logic. While pioneering works have attempted to solve complex legal problems through internalized logical deduction~\citep{HK-O1aw,cai2025unilaw,han2025courtreasoner,cao2025safelawbench}, they face a critical bottleneck: reliance on "closed-loop" reasoning without interaction with external knowledge. 

This isolation results in two primary limitations: 

\textbf{(i) The Introspection Deficit:} LRMs suffer from a fundamental lack of self-awareness regarding their knowledge boundaries. LRMs frequently fail to distinguish between verified knowledge and information gaps, resulting in confident hallucinations rather than acknowledgments of uncertainty~\cite{dahl2024large,blair2025llms}. Our experiment shows that in over 70\% of error cases, models fail to trigger necessary verification even when tools are available. This indicates that the primary failure mode is not merely a lack of knowledge, but a deficit in introspection, which is the inability to decide whether to search.

\textbf{(ii) Fragility in Complex Legal Scenarios:} In our preliminary experiments, we observe that for complex legal reasoning scenarios, relying solely on full search is completely insufficient. We find a substantial performance gap between full search and agentic search on deep reasoning, indicating that static methods struggle to bridge the reasoning deficit. This inherent inflexibility largely restricts the upper bound of the models' abilities. Consequently, models should learn to search autonomously rather than passively processing static search results. This enables them to adapt to diverse and complex legal scenarios.

To address these limitations, we introduce the LRAS (Legal Reasoning with Agentic Search), a framework designed to shift the paradigm of legal reasoning from static, parametric "closed-loop thinking" to dynamic, interactive "Active Inquiry" through dynamic interaction (as shown in Figure~\ref{fig:legal_comparison}). LRAS is the first framework to explicitly couple legal reasoning with agentic search, moving beyond passive retrieval to proactive, iterative information seeking. The framework integrates two core learning mechanisms, each targeting a fundamental weakness of existing methods: \textbf{(i) Introspective Imitation Learning} directly mitigates introspection deficit by training the model to recognize its own knowledge boundaries. By emulating expert query refinement, the model learns to proactively trigger external searches when faced with ambiguous or precision-critical legal content, thereby grounding its responses in verified information. \textbf{(ii) Difficulty-aware Reinforcement Learning} resolves the inherent inflexibility found in complex legal reasoning scenarios. By enabling the model to autonomously plan and execute multi-step exploratory searches, this approach counters the limitations of static methods that fail due to knowledge blind spots. This mechanism evolves the system from the reactive, boundary-triggered search of imitation learning to the autonomous handling of complex legal questions, effectively bridging the reasoning gap.

Our main contributions are summarized as follows:

\begin{itemize}
\item We introduce LRAS, the first framework that shifts legal LLMs from static "closed-loop thinking" to dynamic "Active Inquiry," enabling models to identify knowledge boundaries and handle legal reasoning complexity.

\item We design a dual-mechanism learning architecture, comprising Introspective Imitation Learning and Difficulty-aware Reinforcement Learning. This effectively resolves the introspection deficit regarding whether to search, and overcomes the fragility of static retrieval in complex legal scenarios by autonomously learning how to search.

\item Extensive experiments demonstrate the effectiveness of LRAS, showing performance gains from 8.2\% to 32\%, particularly in tasks requiring deep reasoning with
reliable knowledge.

\end{itemize}

%% file: sections/related_work.tex
\subsection{Legal Reasoning with LLMs}

Early LLM applications in legal domains focused on general-purpose tasks such as judicial judgment prediction and contract generation~\citep{hou2025large,dehghani2025large}. Subsequent benchmarks,(e.g.LegalBench~\citep{guha2023legalbench}, LawBench~\citep{fei2024lawbench}, and LexEval~\citep{li2024lexeval}) revealed significant shortcomings in the legal domain, prompting a shift toward step-by-step reasoning enhancement. Recent efforts have proposed to formalize legal syllogism (LoT~\citep{jiang2023legal} and LawChain~\citep{xie2025lawchain}) or improve reliability via process supervision (LegalReasoner~\citep{shi2025legalreasoner} and SyLeR~\citep{zhang2025syler}),  while agentic frameworks like GLARE~\citep{yang2025glare} seek to deepen legal reasoning by invoking dynamic modules. However, most current approaches operate predominantly as closed-book systems, relying on static internal parameters rather than dynamic external knowledge retrieval. This constraint limits their ability to transition from "closed-loop thinking"—reasoning solely within fixed model weights—to "active inquiry," in which models proactively engage with external information sources. To address this limitation, we propose the first unified framework that integrates agentic search with legal reasoning to enable this paradigm shift, equipping LLMs with the capacity for dynamic interaction with external legal environments. 
 
\subsection{Legal-Aware Reinforcement Learning}

Recent work has adapted reinforcement learning (RL) to enhance legal reasoning in LLMs. One line of research structures logical reasoning through legal‑augmented or structure‑aware rewards, as in SyLeR~\citep{zhang2025syler}, LegalDelta~\citep{dai2025legal}, and Unilaw‑R1~\citep{cai2025unilaw}, to enforce rigorous syllogistic reasoning. Another focuses on factual and procedural correctness, employing strategies such as citation‑aware rewards~\citep{akarajaradwong2025can} and hard‑sample mining (e.g., HIPO~\citep{hu2025fine}) to reduce hallucinations and ensure precision. Further efforts like LexPam~\citep{zhang2025legal} extend RL to legal‑mathematical reasoning via curriculum learning. Despite these advances, current legal RL methods remain limited by a static training environment hindering models from adaptively navigating complex legal scenarios, a gap which our proposed method seeks to bridge.

%% file: sections/pre_and_moti.tex
\subsection{Preliminary}\label{sec:preliminary}

\paragraph{Problem Formulation.}
We formulate legal agentic search as a sequential decision-making process augmented by an external search environment. Let $x \in \mathcal{X}$ denote the input legal query and $\mathcal{E}_{\text{search}}$ represent the external search environment. Unlike standard LRMs that rely solely on parametric memory to map $x \to y$, \ourmodel~operates as a policy $\pi_\theta$. The agent actively interacts with $\mathcal{E}_{\text{search}}$ to acquire external legal information, constructing a reasoning trajectory $\tau$ to support its final answer $y$.

\paragraph{Introspective Action Space.}
The reasoning process unfolds sequentially over $T$ steps. At each step $t$, the agent generates an action $z_t$ conditioned on the input $x$ and the history $h_t = (z_1, o_1, \dots, z_{t-1}, o_{t-1})$. We define the action space $\mathcal{Z}$ as the following three types:

\begin{itemize}[leftmargin=*,noitemsep] 
    \item \textbf{Introspective Reasoning ($z_t^{\text{think}}$):} The process initiates with internal thinking (encapsulated within \texttt{<think>...</think>}). The agent evaluates whether its parametric knowledge is sufficient. Based on this assessment, it transitions to one of two subsequent states.

    \item \textbf{Active Inquiry ($z_t^{\text{search}}$):} If internal knowledge is deemed insufficient, the agent formulates a search query (encapsulated within \texttt{<search>...</search>}). This action triggers the environment to return an observation $o_t = \mathcal{E}_{\text{search}}(z_t)$ (wrapped in \texttt{<information>...</information>}). The process then loops back to Introspective Reasoning to assess the adequacy of the new information.
    
    \item \textbf{Answer Generation ($y$):} Once the agent determines that the accumulated information in history $h_t$ is sufficient, it terminates the search loop and generates the final legal response.
\end{itemize}

Thus, the complete trajectory can be defined as $\tau = (z_1^{\text{think}}, [z_1^{\text{search}}, o_1, z_2^{\text{think}}, \dots], y)$, representing an introspective process of dynamic think and active inquiry that concludes upon reaching knowledge sufficiency.

\paragraph{Objective.}
The objective is to maximize the accuracy of legal answers conditioned on introspective reasoning:
\begin{equation}
    \max_\theta \sum_{i=1}^{N} \mathbb{I}(y_i = y_i^*), \quad y_i = \pi_\theta(x_i, \tau_i),
\end{equation}
where $\tau_i$ is the reasoning trajectory (think and search actions) for query $x_i$, and $y_i$ is the answer generated after trajectory completion.

\subsection{Motivation}\label{sec:motivation}

\input{tables/passrate_0_search}

To characterize the limitations of current LRMs in the legal domain, we conduct exploration experiments and identify a critical bottleneck: LRMs suffer from a fundamental lack of introspection, frequently resulting in confident hallucinations rather than acknowledgments of uncertainty. This deficit is exemplified in our case study (Figure~\ref{fig:app_case_study}), where the model fabricates a "one-year" deadline alongside a spurious rationale, failing to distinguish between verified knowledge and information gaps. Quantitative analysis of 705 incorrect responses (Table \ref{tab:mov_search_analysis}) confirms this pattern: while external search tools could have corrected 8.2\% of errors, the model triggered the search mechanism in only 28.7\% of failure instances. This indicates that in over 71\% of cases, the model failed to recognize its own need for verification. Consequently, the primary failure mode is not a mere lack of knowledge, but a deficit in introspective capability—the inability to identify knowledge boundaries prevents the model from utilizing available tools to remedy its errors. Consequently, this evidence directly motivates our approach of integrating search as a core mechanism to enhance the reliability and extend the functional boundaries of legal LLMs.

\begin{figure}[t]
    \centering
    \includegraphics[width=1.0\linewidth]{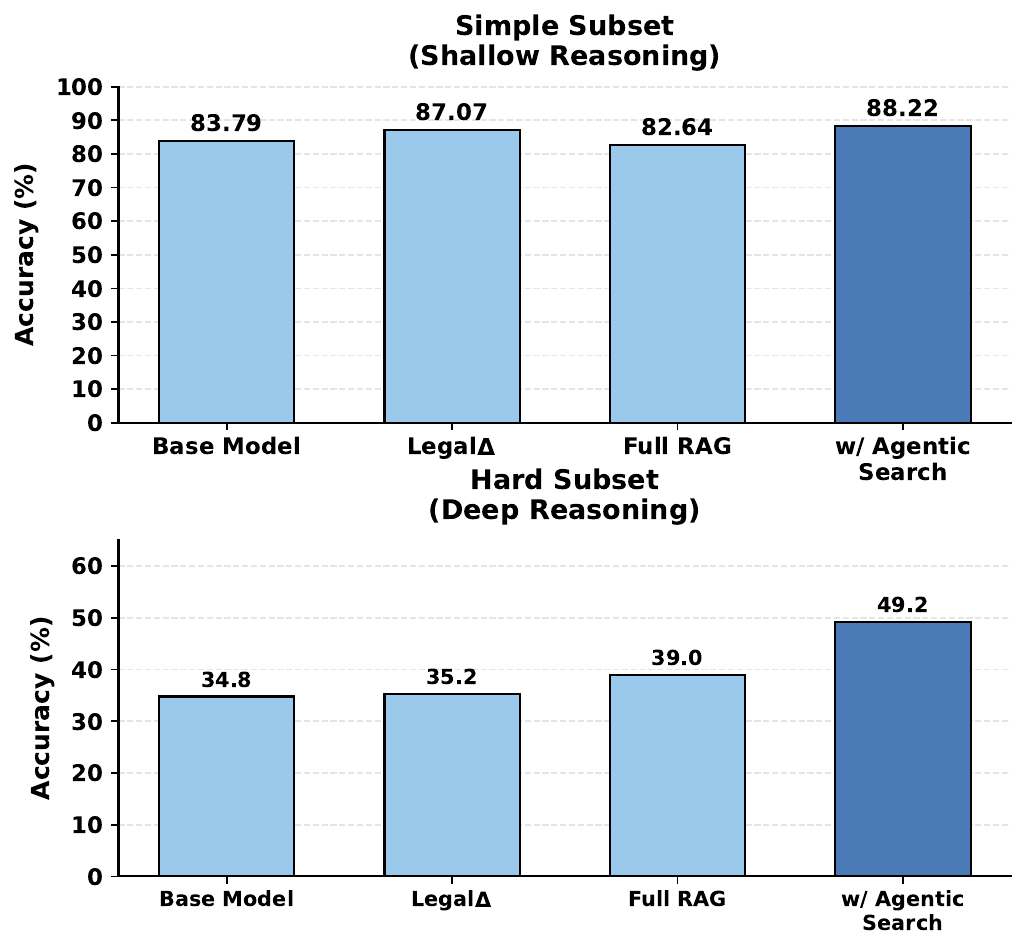} 
    \caption{Impact of search strategies on Shallow and Deep Reasoning.}
    \label{fig:motivation_agentic}
    \vspace{-2mm}
\end{figure}

While external knowledge is critical for complex legal reasoning, standard retrieval methods like Full RAG, which incorporate an external knowledge database with the model for a static pipeline, prove insufficient. As illustrated in Figure \ref{fig:motivation_agentic}, the necessity of search varies significantly by difficulty. On the Simple Subset (Shallow Reasoning), internal knowledge is largely adequate, with search yielding only marginal gains (improving the Base Model from 83.79\% to 88.22\%). In stark contrast, the Hard Subset (Deep Reasoning) reveals a severe capability gap, where models plateau around 35\% accuracy. Crucially, passive retrieval strategies fall short here: standard "Full RAG" only lifts performance to 39.0\%. However, Agentic Search achieves a breakthrough, reaching 49.2\%. This substantial margin demonstrates that complex legal problems require more than just context; they demand an agentic capability to actively navigate and synthesize information.

These observations motivate our \textbf{\ourmodel} framework. To address the introspection deficit, we employ Introspective Imitation Learning, teaching the model to recognize its knowledge boundaries and decide "whether to search" to avoid precision cased by the Introspection Deficit. To further increase the fragility in Complex Legal Scenarios, the model's autonomous decision capacity across complex and underspecified scenarios, we introduce Difficulty-aware Reinforcement Learning, which can teach models "how to search" by shifting from passive context consumption to systematic, evidence-based exploration for complex queries.

%% file: tables/passrate_0_search.tex
\begin{table}[t]
\centering
\small

\renewcommand{\arraystretch}{1.2}
\setlength{\tabcolsep}{8pt} 
\resizebox{\linewidth}{!}{
\begin{tabular}{lr}
\toprule
\textbf{Category} & \textbf{Count (Percentage)} \\
\midrule
\rowcolor{gray!10}
\textbf{Total Base Model Failures} & \textbf{705 (100\%)} \\
\midrule
Samples Where Search Was Triggered & 202 (28.7\%) \\
Samples Where Search Was \textit{Not} Triggered & 503 (71.3\%) \\
\midrule
\rowcolor{green!10}
\textbf{Successfully Corrected by Search} & \textbf{58 (8.2\% of Total)} \\
\textit{Search Success Rate (Corrected / Triggered)} & \textit{28.7\%} \\
\bottomrule
\end{tabular}
}
\caption{Impact of Search Augmentation on Failed Queries. Among 705 instances randomly sampled from JEC-QA where the base model failed, search was triggered in only 28.7\% of cases (revealing a lack of introspection). When triggered, search successfully corrected the answer 28.7\% of the time.}

\label{tab:mov_search_analysis}
\end{table}

%% file: sections/approach.tex
\subsection{Overview}

\begin{figure*}[t]
\centering
\includegraphics[width=1.0\linewidth]{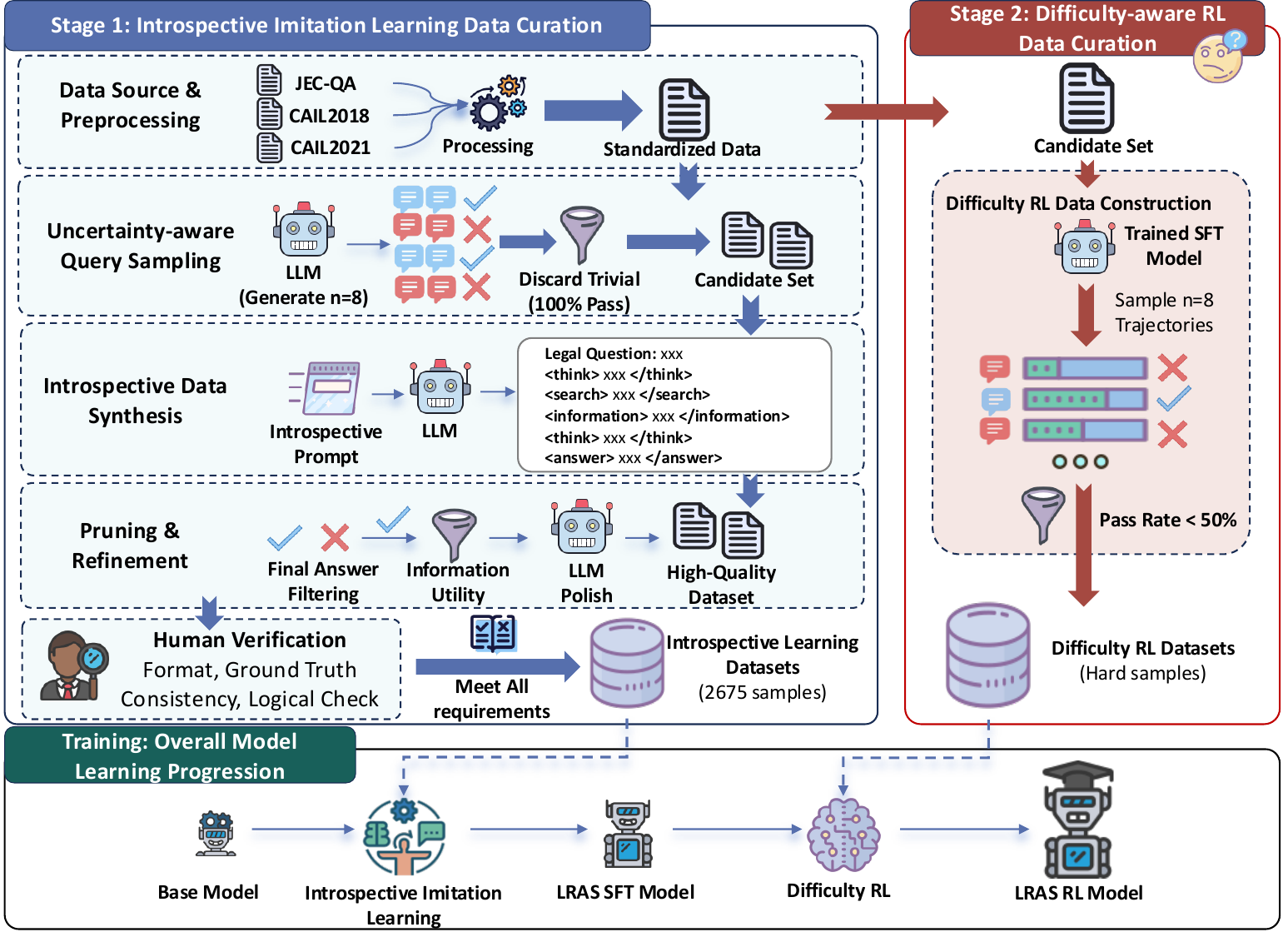}
\caption{The overall framework of our proposed method. The workflow consists of two main phases: Stage 1, Introspective Imitation Learning Data Curation, where raw legal data is processed, filtered for uncertainty, and synthesized into reasoning trajectories with search actions; and Stage 2, Difficulty-aware RL Data Curation, where hard samples are selected based on pass rates to train the model via reinforcement learning. The bottom panel illustrates the progressive training pipeline from the Base Model to the final LRAS RL Model.}
\label{fig:framework}
\end{figure*}

We propose a dual-mechanism learning architecture designed to address two key challenges prevalent in the legal domain: the Introspection Deficit and Fragility in Complex Legal Scenarios. Our framework integrates Introspective Imitation Learning and Difficulty-aware Reinforcement Learning to achieve precise, proactive, and trustworthy legal reasoning as shown in Figure~\ref{fig:framework}.

\vspace{-2mm}

\begin{itemize}[leftmargin=*,noitemsep] 
 \item \textbf{Stage 1: Introspective Imitation Learning.} To mitigate the Introspection Deficit, we establish a specialized data curation pipeline. By employing uncertainty-aware query sampling and introspective process synthesis, we construct a dataset that explicitly models knowledge boundaries. Through introspective learning, we train the model to emulate expert behaviors, proactively triggering external searches only when facing ambiguous or precision-critical legal content.\\
    
\item \textbf{Stage 2: Difficulty-aware Reinforcement Learning.} To counter the fragility in complex, underspecified scenarios, we advance beyond static supervision compared with stage 1. We construct a "hard" dataset based on SFT failure rates and employ Group Relative Policy Optimization (GRPO). This stage empowers the model to autonomously plan and execute multi-step exploratory searches, evolving the agent from reactive fact-checking to the deep reasoning required for navigating intricate legal problems.
\end{itemize}

\subsection{Introspective Imitation Learning}

\subsubsection{Introspective Data Synthesis}
\label{sec:sft_curation}

To facilitate Introspective Imitation Learning, we curate a high-quality dataset that compels the model to learn structured reasoning and search invocation. This process involves a rigorous pipeline of sourcing, synthesis, and verification.

We derive our initial corpus from three primary legal benchmarks: JEC-QA~\citep{zhong2020jec}, CAIL2018~\citep{xiao2018cail2018}, and CAIL2021~\citep{cail2021github}, which provide a mix of objective multiple-choice and complex QA entries. To ensure the training data provides a sufficient learning signal, we implement an uncertainty-aware query sampling strategy. Using Qwen3-235B-Instruct, we generate $n=8$ candidate responses for each source question. We discard trivial queries where the model achieves a perfect pass rate, retaining only those exhibiting inherent difficulty or reasoning ambiguity. This filtering ensures the dataset focuses on scenarios where the model's internal parametric knowledge is insufficient, necessitating external retrieval.

For the retained queries, we synthesize high-quality reasoning trajectories via prompt-based generation. We enforce a structured format using specialized tags: \texttt{<think>} for internal reasoning, \texttt{<search>} for query generation, \texttt{<information>} for retrieval, and \texttt{<answer>} for the final response. To eliminate hallucinations and redundant operations, we apply a two-stage trajectory pruning and refinement process. First, trajectories yielding incorrect answers are discarded. Second, we employ Claude-4.5-Sonnet to evaluate the utility of retrieved information, removing trajectories where search actions fail to support the query (i.e., $\exists I_i \in T, \text{Support}(I_i, Q) = \text{False}$). Remaining high-quality trajectories are rewritten for fluency and logical consistency.

The final stage involves rigorous human verification. A professional annotator reviews the synthesized data for format correctness, ground-truth consistency, and the logical necessity of search operations. This process yields 2,675 high-quality samples, as detailed in Appendix~\ref{app:training_setup}.

\vspace{-2mm}

\input{tables/main_table}

\subsubsection{Training Objective}

We perform full-parameter Supervised Fine-Tuning (SFT) on the Qwen3 series models. Each sample $s$ in the training dataset $\mathcal{S}$ consists of a question $x$ and a structured trajectory $t = (t_1, t_2, \ldots, t_n)$. The model optimizes the standard autoregressive language modeling objective:
\begin{equation}
\mathcal{L}_{\text{SFT}} = -\sum_{i=1}^{n} \log P(t_i | x, t_{<i})
\end{equation}
By minimizing this loss, the model learns to generate complete trajectories with proper tag usage. Crucially, this phase instills the "introspective" capability: the model learns to internalize the expert behavior of recognizing knowledge boundaries and initiating searches only when necessary, rather than hallucinating answers to complex legal queries.

\subsection{Difficulty-aware Reinforcement Learning}

\subsubsection{Difficulty-aware Data Synthesis}
To maximize training efficiency, we construct a dataset specifically focused on hard samples where the supervised model struggles. We utilize the trained SFT model to sample $n=8$ trajectories for each query in the original dataset. We filter these queries based on performance, retaining only those with a pass rate below 50\%. This selection strategy targets complex scenarios where the SFT model, despite invoking the search tool, failed to derive the correct answer. Consequently, the RL phase concentrates on correcting logic gaps and refining search strategies for long-tail knowledge.

\subsubsection{Training Objective}
We employ Group Relative Policy Optimization (GRPO)~\citep{shao2024deepseekmath}. In each training iteration, for a question $x$, we sample $K$ candidate trajectories $\{y_k\}_{k=1}^K$ from the current policy $\pi_\phi$. Each trajectory receives a reward $R_k$. We compute the group-normalized advantage $A_k = (R_k - \bar{R}) / \sigma_R$, where $\bar{R}$ and $\sigma_R$ are the mean and standard deviation of rewards within the group. The policy is updated to maximize:
\begin{equation}
\small
\begin{split}
\mathcal{L}_{\text{GRPO}}(\phi) = \mathbb{E}_{x \sim \mathcal{D}, \{y_k\}_{k=1}^K \sim \pi_\phi} \Bigg[ \frac{1}{K} \sum_{k=1}^K \min \Big( \rho_k A_k, \quad\quad \\
\text{clip}(\rho_k, 1-\epsilon, 1+\epsilon) A_k \Big) \Bigg] - \lambda \text{KL}(\pi_\phi || \pi_{\text{ref}})
\end{split}
\end{equation}
where $\rho_k$ is the probability ratio between the new and old policies, and $\lambda$ controls the KL divergence penalty.

To guide the model toward both accuracy and structural integrity, we employ a composite reward function $R(x, y)$ that evaluates the final answer and format compliance:
\begin{equation}
\small
R(x, y) =
\begin{cases}
1 & \text{if } a_{\text{pred}} = a_{\text{gold}} \land f_{\text{fmt}}(y), \\
1 - \lambda_f & \text{if } a_{\text{pred}} = a_{\text{gold}} \land \neg f_{\text{fmt}}(y), \\
\lambda_f & \text{if } a_{\text{pred}} \neq a_{\text{gold}} \land f_{\text{fmt}}(y), \\
0 & \text{otherwise},
\end{cases}
\end{equation}
where $f_{\text{fmt}}(y)$ validates the correct usage of tags (e.g., \texttt{<think>}). This reward structure incentivizes the model to produce correct answers while strictly adhering to the required output format.

%% file: tables/main_table.tex
\begin{table*}[!t]
\centering
\renewcommand{\arraystretch}{1.05}
\setlength{\tabcolsep}{8pt}
\definecolor{ourscolor}{RGB}{245, 245, 245}
\small
\begin{tabular}{l c c c c c c}
\toprule
\textbf{Model} & \textbf{Parameters} & \textbf{LexEval} & \textbf{LawBench} & \textbf{Unilaw-R1-Eval} & \textbf{DiscLaw} & \textbf{Avg.(\%)} \\
\midrule
\multicolumn{7}{c}{\textit{Legal LLMs}} \\
\cdashline{1-7}
Fuzi-7B             & 7B  & 13.56 & 41.22 & 6.50 & 17.55 & 19.71 \\
LawLLM-7B         & 7B & 49.72 & 77.87 & 29.12 & 48.25 & 51.24 \\
DiscLaw-13B         & 13B & 35.07 & 59.49 & 22.38 & 33.11 & 37.51 \\
Legal$\Delta$-3B    & 3B  & 33.58 & 71.39 & 26.50 & 51.56 & 45.76 \\
Legal$\Delta$-7B    & 7B  & 48.02 & 75.78 & 37.88 & 45.75 & 51.86 \\
Legal$\Delta$-14B   & 14B & \textbf{49.74} & \textbf{80.42} & \textbf{45.88} & \textbf{52.14} & \textbf{57.05} \\
\midrule
\multicolumn{7}{c}{\textit{Qwen3-4B}} \\
\cdashline{1-7}
Base                              & 4B & 38.96 & 61.67 & 30.50 & 49.47 & 45.15 \\
\rowcolor{ourscolor} \ourmodel-SFT        & 4B & 51.37 & 72.38 & 38.25 & 61.64 & 55.91\textcolor{red}{$\uparrow$23.8\%} \\
\rowcolor{ourscolor} \ourmodel-RL   & 4B & \textbf{51.40} & \textbf{76.83} & \textbf{44.38} & \textbf{65.72} & \textbf{59.58}\textcolor{red}{$\uparrow$32.0\%} \\
\midrule
\multicolumn{7}{c}{\textit{Qwen3-8B}} \\
\cdashline{1-7}
Base                              & 8B & 49.22 & 71.96 & 39.50 & 56.93 & 54.40 \\
\rowcolor{ourscolor} \ourmodel-SFT       & 8B & 55.29 & 75.94 & 49.38 & 67.18 & 61.95\textcolor{red}{$\uparrow$13.9\%} \\
\rowcolor{ourscolor} \ourmodel-RL   & 8B & \textbf{57.44} & \textbf{78.76} & \textbf{51.00} & \textbf{73.23} & \textbf{65.11}\textcolor{red}{$\uparrow$19.7\%} \\
\midrule
\multicolumn{7}{c}{\textit{Qwen3-14B}} \\
\cdashline{1-7}
Base                              & 14B & 52.28 & 73.88 & 43.00 & 66.59 & 58.94 \\
\rowcolor{ourscolor} \ourmodel-SFT       & 14B & 57.07 & 78.76 & 49.38 & 71.20 & 63.80\textcolor{red}{$\uparrow$8.2\%} \\
\rowcolor{ourscolor} \ourmodel-RL   & 14B & \textbf{59.84} & \textbf{80.10} & \textbf{54.37} & \textbf{75.66} & \textbf{67.49}\textcolor{red}{$\uparrow$14.5\%} \\
\bottomrule
\end{tabular}
\caption{Evaluation results across different benchmarks and model sizes. \colorbox{ourscolor}{Highlighted rows} indicate our models. \textcolor{red}{Red arrows} denote relative percentage improvement over the corresponding base model.}
\label{tab:main_results}
\vspace{-2mm}
\end{table*}

%% file: sections/experiments.tex
\subsection{Experimental Setup}

\paragraph{Benchmarks and Metrics.}
We conduct evaluations on three in-distribution Chinese legal domain multi-task benchmarks: LexEval~\citep{li2024lexeval}, LawBench~\citep{fei2024lawbench}, and UniLaw~\citep{cai2025unilaw}, along with one out-of-distribution (OOD) benchmark, DiscLaw~\citep{yue2024lawllm}. The detailed task descriptions are provided in Appendix~\ref{appendix_datasets}. For evaluation metrics, we follow the original metrics specified by each benchmark for their respective tasks.

\paragraph{Baselines.}
We evaluate \ourmodel-SFT and \ourmodel-RL against representative baselines from two categories. First, we compare with existing legal language models: Fuzi-7B~\citep{sdu2023fuzimingcha}, LawLLM-7B~\citep{yue2024lawllm}, and DiscLaw-13B~\citep{yue2023disc}. Second, we include specialized legal reasoning models: Legal$\Delta$-3B/7B/14B~\citep{dai2025legal}. We also report results for the base model to demonstrate the effectiveness of our legal domain adaptation.

\paragraph{Implementation Details.}
We adopt the Qwen3 series (4B/8B/14B) as backbone models to ensure comprehensive evaluation across different model scales. For legal-aware imitation learning, we perform full-parameter fine-tuning with a learning rate of $1 \times 10^{-5}$, using cosine learning rate scheduling with 10\% warmup ratio over 3 epochs. For difficulty-aware reinforcement learning, we employ the GRPO advantage estimator without KL penalty in the reward function. The actor model is optimized with a learning rate of $1 \times 10^{-6}$, while the critic uses $1 \times 10^{-5}$. We set the rollout sample size to 5 per prompt and train for 15 epochs. Additional hyperparameters are provided in Appendix~\ref{app:hyperparams}.

\begin{figure}[t]
\centering
\includegraphics[width=1.0\linewidth]{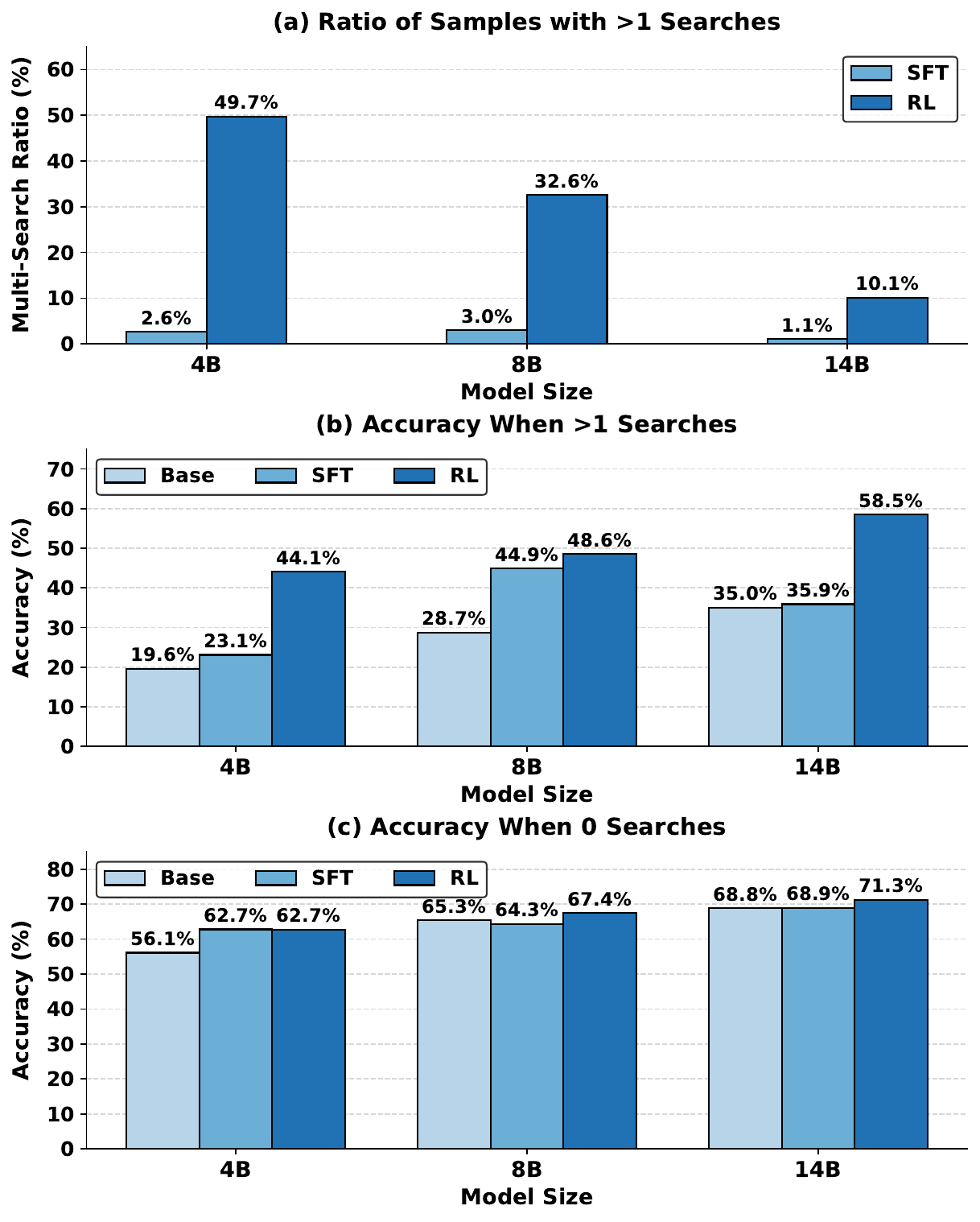}
\caption{Analysis of search behaviors across model sizes. (a) Proportion of samples with $>1$ search actions. (b) Accuracy on samples requiring multi-step search. (c) Accuracy on samples with 0 searches.}
\label{fig:multi-search}
\vspace{-3mm}

\end{figure}

\begin{figure*}[t]
    \centering
    \includegraphics[width=0.85\textwidth]{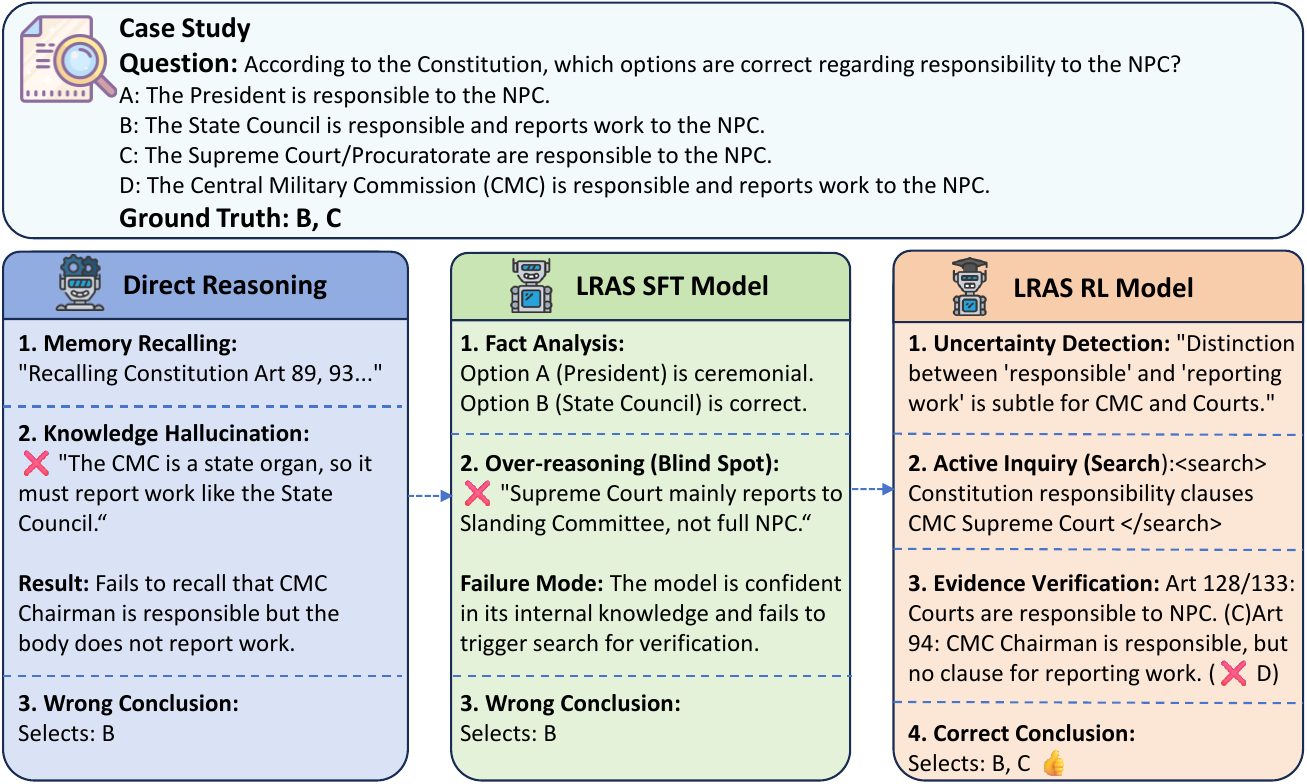}
    \caption{A case study comparing reasoning processes.}
    \label{fig:case_study}
    \vspace{-3mm}
\end{figure*}

\subsection{Main Results and Findings}

\noindent\textbf{\ourmodel~achieves state-of-the-art performance across all benchmarks, substantially outperforming both pure reasoning models and existing legal domain models.}
Table~\ref{tab:main_results} presents the evaluation results of \ourmodel~across multiple benchmarks. \ourmodel-RL (14B) attains an average score of 67.49\%, surpassing the previous best model Legal$\Delta$-14B by 18.3\% relatively. The progression from base models to \ourmodel-SFT to \ourmodel-RL demonstrates the cumulative effectiveness of both training stages. Specifically, Introspective imitation learning teaches models to identify knowledge boundaries and initiate searches appropriately, yielding substantial improvements over base models (e.g., 8.2\% relative improvement for 14B). Subsequently, difficulty-aware RL enables models to conduct autonomous multi-round exploratory search for complex scenarios which need deep reasoning, providing additional relative gains of 5.8\% on average. Notably, even our smallest variant \ourmodel-RL (4B) achieves 59.58\%, outperforming the larger Legal$\Delta$-14B (57.05\%) by 4.4\% relatively, demonstrating that our agentic search paradigm—shifting from closed-loop thinking to active inquiry—is more effective than scaling model parameters alone. Furthermore, consistent improvements across all model sizes (4B, 8B, 14B) validate the robustness of our approach.

\noindent\textbf{LARS demonstrates strong generalization capability, maintaining robust performance on OOD legal scenarios.}
\ourmodel-RL (14B) achieves 75.66\% on the DiscLaw benchmark, representing a 13.6\% relative improvement over the base model and outperforming all baseline legal models. Importantly, both training stages contribute to enhanced OOD performance: introspective imitation learning improves DiscLaw scores by 6.9\% relatively, while difficulty-aware RL adds another 6.3\%. This indicates that learning to actively inquire and verify information—rather than relying solely on parametric memory—fundamentally strengthens the model's ability to handle diverse legal scenarios. The consistent gains across both in-distribution benchmarks (LexEval, LawBench, and Unilaw-R1-Eval) and OOD DiscLaw suggest that our approach develops generalizable legal reasoning capabilities.

\noindent\textbf{Difficulty-aware RL fundamentally stimulates \ourmodel's active inquiry capability, transforming it from a passive responder into an adaptive investigator.} As shown in Figure \ref{fig:multi-search}(a), SFT models rarely engage in deep exploration (<3\% multi-search rate), whereas RL drastically increases this behavior (e.g., 49.7\% for 4B); At the same time, Figure \ref{fig:multi-search}(b) confirms that this active inquiry directly translates to performance gains on complex queries, with the RL-14B model outperforming SFT by over 22\% (58.5\% vs. 35.9\%). Furthermore, Figure \ref{fig:multi-search}(c) demonstrates that this aggressive search strategy does not compromise performance on simple, knowledge-internal queries ($0$ searches), where the RL model maintains or slightly exceeds the high accuracy of the SFT baseline.

\noindent\textbf{Ablation Studies of the Introspection Levels.} To investigate the effectiveness of our hierarchical reasoning design, we conduct an ablation study across three progressive levels of introspection. As shown in Table \ref{tab:introspection_ablation}, increasing the depth of self-reflection consistently yields performance gains across all benchmarks. Specifically, the full \textbf{Level-3} configuration achieves the highest average score of \textbf{60.20\%}, significantly outperforming the baseline Level-1 by a margin of \textbf{+4.16\%}. This trend is particularly pronounced on the more complex UniLaw dataset (\textbf{+7.88\%}), validating that deeper introspection is crucial for handling intricate legal scenarios.

\subsection{Case Study}

Figure~\ref{fig:case_study} illustrates a case study comparing model performance on a constitutional law query. Both baseline models failed: the Direct Reasoning model hallucinated non-existent reporting duties for the Central Military Commission, while LRAS-SFT suffered from overconfidence, failing to verify its internal knowledge. Conversely, LRAS-RL succeeded by recognizing the query's complexity. It utilized Active Inquiry to retrieve specific legal articles, allowing it to correctly distinguish between the concepts of "responsibility" and "reporting work," thereby selecting the correct answer.

%% file: sections/conclusion.tex
In this work, we introduce \ourmodel, a framework designed to shift the paradigm of legal reasoning from static "closed-loop thinking" to dynamic "Active Inquiry." By integrating Introspective Imitation Learning with Difficulty-aware Reinforcement Learning, \ourmodel~effectively addresses the fundamental introspection deficit by teaching models whether to search, while simultaneously overcoming the fragility in complex scenarios by enabling them to learn how to search autonomously. Extensive experiments demonstrate that our approach significantly outperforms static retrieval methods, successfully bridging the reasoning deficit in deep legal reasoning tasks. Ultimately, we believe LRAS establishes a robust paradigm for autonomous legal agents, contributing meaningfully to the advancement of trustworthy and precise Legal AI.

%% file: sections/limitations.tex
Despite notable advancements, LRAS faces several limitations. Due to computational resource constraints, we limited the models size and types of tools. Consequently, the model may not fully explore extremely complex scenarios requiring deep, multi-layered inquiry chains. Future work will explore extended interaction horizons to handle broader long-tail complexities. Additionally, LARS’s active inquiry relies heavily on the precision of external tools. If the retrieval system returns incomplete or irrelevant statutes, the model's subsequent reasoning may be compromised. We will integrate robustness mechanisms to better detect and correct tool-induced errors in our future work.

%% file: sections/appendix.tex
\section{Datasets}\label{appendix_datasets}

\subsection{Lexeval}

LexEval is a comprehensive Chinese legal benchmark comprising 23 tasks and approximately 14,150 questions. It introduces the Legal Cognitive Ability Taxonomy (LexCog) to organize tasks systematically. The data is derived from existing legal datasets, real-world examinations, and expert annotations. LexEval evaluates models not only on fundamental legal knowledge application but also on their ability to handle ethical issues within legal contexts.

\subsection{Lawbench}

LawBench is designed to evaluate legal capabilities across three specific cognitive levels: (1) Legal Knowledge Memorization (recall of concepts, articles, and facts); (2) Legal Knowledge Understanding (comprehension of entities and relationships); and (3) Legal Knowledge Application (reasoning to solve realistic tasks). It contains 20 diverse tasks categorized into single-label classification, multi-label classification, regression, extraction, and generation.

\subsection{UniLaw-Eval}

UniLaw-R1-Evalconsists of 800 items, split into 426 single-choice and 374 multi-choice questions. It is designed to rigorously test the model's ability to perform logical inference within legal scenarios.

\subsection{Disc-Law}

DISC-Law (also referred to as Law-Eval) assesses legal LLMs from both objective and subjective dimensions: Objective evaluation consists of multiple-choice questions sourced from standardized examinations (e.g., National Judicial Examination, CPA). Questions are categorized into Easy, Normal, and Hard levels to test knowledge retrieval and deduction. Subjective evaluation comprises 300 manually constructed examples covering legal consultation and judgment prediction. Evaluation utilizes a referee model (e.g., GPT-3.5) to score responses based on accuracy, completeness, and clarity.

\subsection{Selected Tasks}

We focus on knowledge-intensive tasks that require legal knowledge retrieval and reasoning. Specifically, we evaluate on:
\begin{itemize}
    \item \textbf{LexEval}: Tasks 1-1, 1-3, 3-1, 3-2, 3-3, 3-4, 3-5, 3-6, 4-1, and 4-2
    \item \textbf{LawBench}: Tasks 1-2, 3-1, 3-3, 3-4, 3-5, and 3-6
    \item \textbf{UniLaw-Eval}: Complete benchmark (all 800 questions)
    \item \textbf{DISC-Law}: Complete benchmark (both objective and subjective evaluation sets)
\end{itemize}

\section{Dataset Licenses}

Several foundational Chinese legal datasets are publicly available under open-source licenses, including JEC-QA, CAIL2018, and CAIL2021. Additionally, LexEval is released under the MIT license, while both LawBench and DISC-LawLLM are available under the Apache-2.0 license.

\section{Ablation Results}

The ablation results on introspection levels are presented in Table~\ref{tab:introspection_ablation}.

\input{tables/ablation_table}

\section{Details of the Training Data}\label{app:sft_stats}

\begin{figure}[t]
\centering
\includegraphics[width=1.0\linewidth]{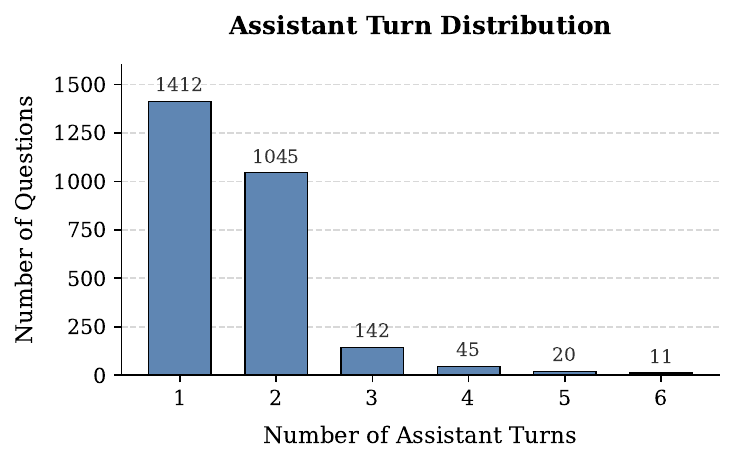}
\caption{Distribution of assistant turns in the SFT dataset, illustrating the frequency of conversation lengths.}
\label{fig:assistant_turn}
\end{figure}

\input{tables/sft_statis}

This section presents the statistical characteristics of the Legal Search-R1 SFT dataset, which contains 2,675 samples. As illustrated in Figure~\ref{fig:turn_distribution}, the dataset exhibits a diverse distribution of conversation lengths, ranging from single-turn queries to complex interactions of up to 6 turns, with a mean of 1.60 turns. Table~\ref{tab:sft_stats} further details these metrics: the reasoning process (encapsulated in \texttt{<think>} tags) averages 703 characters, reaching a maximum of 1,537, indicating substantial cognitive depth. Additionally, the inclusion of search calls (mean=1.27, max=5) in search-based samples supports the robust training of agentic behaviors, while the total response length averages 615 characters.

\section{Details of Training Setup}\label{app:training_setup}

In this section, we provide the detailed hyperparameters and configurations used for training Legal Search-R1. We utilize the Qwen3 series (4B/8B/14B) as our backbone models. All experiments are conducted on a cluster of 8 NVIDIA GPUs.

\subsection{Hyperparameters}\label{app:hyperparams}

Table~\ref{tab:hyperparameters} lists the specific hyperparameters used for both the Supervised Fine-Tuning (SFT) and Reinforcement Learning (RL) stages. 

For the SFT stage, we employ DeepSpeed ZeRO-3 optimization to handle the full-parameter fine-tuning efficiently. We set the maximum sequence length to 32,768 tokens to accommodate long legal contexts.

For the RL stage, we utilize the GRPO algorithm. A key distinction in our setup is the separate learning rates for the actor ($1 \times 10^{-6}$) and the critic ($1 \times 10^{-5}$) to stabilize training. The reward function is a weighted sum of accuracy and formatting constraints: $R = 1.0 \times \text{Score}_{\text{acc}} + 0.2 \times \text{Score}_{\text{struct}} + 0.1 \times \text{Score}_{\text{format}}$.

\begin{table}[h]
    \centering
    \small
    \begin{tabular}{l|c}
    \toprule
    \textbf{Hyperparameter} & \textbf{Value} \\
    \midrule
    \multicolumn{2}{c}{\textit{Supervised Fine-Tuning (SFT)}} \\
    \midrule
    Optimizer & AdamW \\
    Learning Rate & $1 \times 10^{-5}$ \\
    LR Scheduler & Cosine \\
    Warmup Ratio & 0.1 \\
    Batch Size (Global) & 128 \\
    Epochs & 3 \\
    Max Sequence Length & 32,768 \\
    Precision & bf16 \\
    DeepSpeed Stage & ZeRO-3 \\
    \midrule
    \multicolumn{2}{c}{\textit{Reinforcement Learning (RL)}} \\
    \midrule
    Algorithm & GRPO \\
    Actor Learning Rate & $1 \times 10^{-6}$ \\
    Critic Learning Rate & $1 \times 10^{-5}$ \\
    KL Penalty in Reward & False \\
    Actor KL Loss Coefficient & 0.001 \\
    Rollout Sample Size ($N$) & 5 \\
    Max Prompt Length & 8,192 \\
    Max Response Length & 8,192 \\
    Total Epochs & 15 \\
    Global Train Batch Size & 256 \\
    \bottomrule
    \end{tabular}
    \caption{Hyperparameters for SFT and RL training stages.}
    \label{tab:hyperparameters}
\end{table}

\subsection{Tool Configuration}

The agent is equipped with a legal search tool powered by an external retrieval service (SerpAPI combined with Jina Reader). The configuration for the multi-turn interaction is as follows:
\begin{itemize}
    \item \textbf{Max Assistant Turns:} 6 (aligned with the maximum turns in the SFT dataset).
    \item \textbf{Retrieval Top-K:} 5 results per query.
    \item \textbf{Timeout:} 600 seconds per request to ensure stability during extensive legal database queries.
\end{itemize}

\section{Implementation Details}\label{sec:appendix_search_inference}

\subsection{Search and Summarization Pipeline}

To equip the model with up-to-date legal knowledge, we implement a robust retrieval pipeline consisting of three stages: search, extraction, and summarization.

\begin{enumerate}
    \item \textbf{Search and Extraction:} We utilize \textbf{SerpAPI} to query the internet, retrieving the top-5 most relevant search results for each query generated by the agent. To process the raw URLs, we employ \textbf{Jina Reader}, which effectively extracts clean text content from the webpages, filtering out HTML boilerplate and navigation elements.
    
    \item \textbf{Summarization:} Since raw web content often contains redundancy and noise, we introduce a summarization step before feeding information back to the agent. Following previous works, we deploy an external \textbf{Qwen3-32B} model as a dedicated summarizer. This model condenses the extracted content into concise, query-focused legal abstracts.
    
    \item \textbf{Summarizer Configuration:} For the Qwen3-32B summarizer, we adhere to the official best practices to balance diversity and coherence. We disable the ``thinking'' mode (i.e., \texttt{enable\_thinking=False}) and use the following generation hyperparameters: temperature: $0.7$, top-p: $0.8$, top-k: $20$, min-p: $0$, and a maximum of $8{,}192$ new tokens.
\end{enumerate}

\subsection{Inference Settings for Main Models}

For the evaluation of our trained Legal Search-R1 models (including the backbone Qwen3 series), we prioritize determinism and reproducibility. Therefore, during the inference stage:
\begin{itemize}
    \item We use \textbf{Greedy Decoding} by setting the Temperature to $0$.
    \item The maximum generation length is set to $8,192$ tokens to allow for comprehensive legal reasoning and tool usage.
\end{itemize}

\clearpage
\onecolumn

\section{Prompts}\label{app:prompts}

\input{sections/prompts}


\section{Case Study}

\begin{figure*}[t]
\begin{casestudybox}[width=\textwidth]{Case Study}
\textbf{Question:} When a foreign investor establishes a foreign-funded enterprise, the capital contribution may be paid in installments, but the final installment shall be paid in full within (\underline{\hspace{1cm}}) from the date of issuance of the business license.\\
A: One year \quad B: Two years \quad C: Three years \quad D: Five years \\[0.5em]
\textbf{Ground Truth:} C (Three years)

\tcbline

\textbf{Base Model (Incorrect):}
\begin{quote}
\textit{<think>} According to Article 20 of the "Law of the People's Republic of China on Foreign-funded Enterprises," when establishing a foreign-funded enterprise, the foreign investor shall pay the capital contribution in full within a certain period from the date of issuance of the business license. Installment payment may be allowed, but the deadline for the final installment may be \textcolor{red}{\textbf{one year}}. \textit{</think>}
\end{quote}
\textbf{Prediction: A (One year)} 

\tcbline

\textbf{Agentic Search (Correct):}
\begin{quote}
\textit{<think>} This question examines the regulations on capital contribution deadlines when foreign investors establish foreign-funded enterprises. I need to search for the relevant legal provisions to determine the accurate deadline. \textit{</think>}

\textit{<search>} Foreign investor foreign-funded enterprise installment capital contribution deadline \textit{</search>}

\textit{<information>} ... Article 30 of the "Detailed Rules for the Implementation of the Law of the People's Republic of China on Foreign-funded Enterprises" stipulates that foreign investors may pay capital contributions in installments, but the final installment shall be paid in full within \textcolor{teal}{\textbf{three years}} from the date of issuance of the business license ... \textit{</information>}

\textit{<answer>} C \textit{</answer>}
\end{quote}
\textbf{Prediction: C (Three years)} 
\end{casestudybox}
\caption{Case study of inherent imprecision} 
\label{fig:app_case_study}
\end{figure*}

%% file: tables/ablation_table.tex
\begin{table}[t]
\centering
\small
\renewcommand{\arraystretch}{1.1}
\setlength{\tabcolsep}{4pt} 
\begin{tabular}{lcccc}
\toprule
\textbf{Method} & \textbf{UniLaw} & \textbf{LexEval} & \textbf{LawBench} & \textbf{Avg.} \\
\midrule
\rowcolor{blue!5}
Level-1 & 41.50 & 52.03 & 74.60 & 56.04 \\
\rowcolor{blue!8}
Level-2 & 43.00 & 53.40 & 75.08 & 57.16 \\
\rowcolor{blue!12}
\textbf{Level-3} & \textbf{49.38} & \textbf{55.29} & \textbf{75.94} & \textbf{60.20} \\
\midrule
\rowcolor{green!10}
\textit{$\Delta$ (L3 - L1)} & \cellcolor{green!25}\textit{+7.88} & \textit{+3.26} & \textit{+1.34} & \cellcolor{green!20}\textit{+4.16} \\
\bottomrule
\end{tabular}
\caption{Ablation on introspection levels.}
\label{tab:introspection_ablation}
\end{table}

%% file: tables/sft_statis.tex
\begin{table}[t]
    \centering
    \small
    \renewcommand{\arraystretch}{1.2}
    \setlength{\tabcolsep}{10pt}
    \begin{tabular}{lrr}
        \toprule
        \textbf{Metric} & \textbf{Mean} & \textbf{Max} \\
        \midrule
        \multicolumn{3}{l}{\textit{Interaction Dynamics}} \\
        Conversation Turns & 1.60 & 6 \\
        Search Calls (per question) & 1.27 & 5 \\
        \midrule
        \multicolumn{3}{l}{\textit{Length Characteristics (Chars)}} \\
        Total Response & 615 & 1,512 \\
        Reasoning Process (\texttt{<think>}) & 703 & 1,537 \\
        Search Query & 19.6 & 76 \\
        \bottomrule
    \end{tabular}
    \caption{Statistics of the Legal Search-R1 SFT dataset. Search statistics are computed on search-based samples.}
    \label{tab:sft_stats}
\end{table}

%% file: sections/prompts.tex
This section details the specific prompts designed for our pipeline, categorized into data construction, data rewriting, and reinforcement learning. For the data construction phase, we implemented a three-level introspection mechanism to progressively enhance the model's capability to recognize its knowledge boundaries.

\subsection{Data Construction Prompts}

Our introspection mechanism is divided into three levels, ranging from implicit tool usage to deep metacognitive evaluation.

\begin{figure*}[htbp]
\begin{promptbox}{Level 1}
UNIFIED_SYSTEM = """You are a professional legal assistant. Your task is to accurately answer legal questions.

You possess a search engine tool. When you encounter legal articles, cases, or facts that need verification, you can use the search function.
Your Action Guide:
1. If you believe external information is needed to assist your answer, please directly output <search>Search Query</search>.
2. The system will return <information>Search Results</information>.
3. Combine the search results or your existing knowledge to output <answer>Your Answer</answer>.

Please begin answering directly.
"""
\end{promptbox}
\end{figure*}

\FloatBarrier

\begin{figure*}[htbp]
\begin{promptbox}{Level 2}
UNIFIED_SYSTEM = """You are a professional legal assistant. Your task is to accurately answer legal questions.

You must first write your reasoning process within the <think>Reasoning Process</think> tag. After reasoning, judge whether you lack legal knowledge, and then choose one of the following actions:

**Action 1: Use search tool to find legal information**
If you need to find legal articles, cases, or professional materials to support your answer, use the search tool after thinking. Search results will be returned between <information> Search Summary Content </information>; you may search multiple times as needed.

**Action 2: Provide answer directly**
If external knowledge is no longer required, provide the answer directly after thinking.

### Output Format

**Search Action Format:**
<think>Reasoning Process</think>
<search>Search Query</search>

**Provide Answer Format:**
<think>Reasoning Process</think>
<answer>Your Answer</answer>

### Examples

**Search Action Example:**
<think>
This question involves regulations regarding changes in insurance company shareholders. Option D mentions that "changing contributors or shareholders holding more than 5\% of the company's shares" requires approval. This should be explicitly stipulated in the "Administrative Measures for Equity of Insurance Companies" or relevant laws. I need to find specific articles to confirm.
</think>
<search>Approval for change of shareholders holding more than 5

**Provide Answer Example:**
<think>
According to the search results, Article 16 of the "Administrative Measures for Equity of Insurance Companies" explicitly stipulates: Changing shareholders who hold more than 5\% of the company's shares shall be approved by the insurance regulatory institution. The statement in Option D is consistent with the legal provision.
</think>
<answer>D</answer>

### Key Requirements
1. **The <think> tag must contain substantive reasoning content** and cannot be empty.
2. **Output only one <search> or one <answer> at a time**; do not output both simultaneously.
3. **Do not fabricate the <information> tag yourself**; search results will be automatically provided by the system.
"""
\end{promptbox}
\end{figure*}

\FloatBarrier

\begin{figure*}[htbp]
\begin{promptbox}{Level 3}
UNIFIED_SYSTEM = """You are a professional legal assistant capable of efficiently solving knowledge-intensive tasks. You will answer legal questions.

Note: These questions are of a certain difficulty and require external knowledge to assist reasoning.

**Core Principle - Search Strategy**:
You can utilize your own knowledge or call an external search engine to collect additional information, but searching should only be performed when necessary. Specifically, you should only search when you encounter obvious knowledge gaps or uncertainties that prevent you from answering the question confidently.

**Judgment Criteria**:
Please judge whether to search based on your understanding of the question and your own knowledge reserve. The key is to evaluate:
- Do you have enough certainty and confidence in the answer?
- Will the lack of search affect the accuracy or completeness of the answer?
- Can searching bring substantial information gain?

Prioritize relying on your internal knowledge to answer. Only conduct a search when you clearly realize there is a knowledge gap and this gap will significantly affect the quality of the answer.

To derive the answer, you will proceed step-by-step according to a structured loop.

**Important: You must strictly use the following tag format for output, do not use any other format:**
- `<think>Thinking Content</think>` - Used for thinking and analysis
- `<search>Search Query</search>` - Used for searching (optional)
- `<answer>Final Answer</answer>` - Used to provide the answer

**Thinking Phase (<think>)**: You must first perform the following within the `<think>` tag:
1. Analyze the core requirements of the question.
2. Recall relevant legal knowledge you have mastered.
3. Evaluate whether your knowledge is sufficient to answer confidently.
4. Clearly judge whether a search is needed and explain the reason.
5. If knowledge is sufficient, proceed directly to the answering phase.

**Search Phase (<search>)**: Only construct a precise search query within the `<search>` tag after confirming the existence of a knowledge gap or uncertainty during thinking. If not needed, skip this phase.

**Information Phase (<information>)**: If you performed a search, the results will be returned within the `<information>` tag.

**Answering Phase (<answer>)**: When you have sufficient knowledge, provide a concise and accurate answer within the `<answer>` tag.

Here are a few examples:

---
**Example 1: Search Required**
<think>[Analysis and Judgment]</think>
<search>[Search Query]</search>
<information>[Search Results]</information>
<think>[Continue Analysis]</think>
<answer>C</answer>

---
**Example 2: No Search Required**
<think>[Analysis and Judgment]</think>
<answer>B</answer>

---
**Example 3: Multiple Searches**
<think>[Analysis and Judgment]</think>
<search>[First Search]</search>
<information>[Search Results]</information>
<think>[Continue Analysis]</think>
<search>[Second Search]</search>
<information>[Search Results]</information>
<think>[Final Analysis]</think>
<answer>ACD</answer>

---
**Key Tips**:
- You can search from 0 to N times, deciding autonomously based on actual needs.
- Each search should focus on a clear knowledge gap.
- Final Answer Format: <answer>Answer</answer>
"""
\end{promptbox}
\end{figure*}

\FloatBarrier

\subsection{Data Rewrite Prompts}

This section includes prompts for assessing information relevance and refining the reasoning trajectories.

\begin{figure*}[htbp]
\begin{promptbox}{Relevance Assessment Prompt}
prompt = f"""You are a legal QA quality assessment expert. Please judge whether the following retrieved information can help answer the legal question.

[Judgment Standards]
The information must contain clear legal articles, judicial interpretations, or authoritative definitions that can directly support the key arguments of the answer.

[Specific Requirements]
Must have a clear legal basis.

[Assessment Principles]
- If the information is related to the question topic and provides valuable legal knowledge or basis, judge as YES.
- Only judge as NO if the information is completely off-topic, vague and useless, or unrelated to the question.
- If the information is not perfect but helpful for answering, judge as YES.

Please answer only YES or NO (do not add any explanation):

[Question]
{question}

[Reference Answer]
{final_answer}

[Retrieved Information]
{info_text}

Judgment Result (YES/NO):"""
\end{promptbox}
\end{figure*}

\FloatBarrier

\begin{figure*}[htbp]
\begin{promptbox}{Refinement Prompt (Single Turn)}
rewrite_prompt = f"""You are a professional legal AI assistant training data optimization expert. Please rewrite the analysis process (within the <think> tag) for the following legal question.

Requirements:
- Remove question type priors (do not mention "single choice/multiple choice", etc.).
- Reasoning should be natural, professional, and avoid repetitive legal articles.
- Retain <think></think> and <answer></answer>.
- Do not output <search> or <information>.
- The final answer must be exactly consistent: {original_answer}

Question:
{question}

Original Response:
{assistant_content}

Please output the optimized complete assistant response (containing only think and answer):"""
\end{promptbox}
\end{figure*}

\FloatBarrier

\begin{figure*}[htbp]
\begin{promptbox}{Refinement Prompt (Multi-Turn)}
rewrite_prompt = f"""You are a professional legal AI assistant training data optimization expert. Please rewrite the analysis trajectory for the following legal question.

Hard Requirements:
1) Remove question type priors; do not mention "single choice/multiple choice", etc.
2) Output format allows only: <think>, <search>, <answer>.
3) Do not output <information>.
4) **Must output exactly {expected_search_count} times <search>, writing only the query text each time.**
5) The final answer must be exactly consistent with the original answer: {original_answer}

Question:
{question}

Available Retrieval Evidence (Total {expected_search_count} searches):
{evidence_text}

Please output the optimized complete assistant trajectory (must include {expected_search_count} times <search>, corresponding sequentially to the evidence above):"""
\end{promptbox}
\end{figure*}

\FloatBarrier

\subsection{RL Prompts}

This section details the prompts used during the Reinforcement Learning stage, including the RL prompt and the summarization prompt.

\begin{figure*}[htbp]
\begin{promptbox}{RL Prompt}
USER_CONTENT_PREFIX = """You are a professional legal assistant, skilled in answering legal questions accurately through deep search.

Work Style:
1. First reason and analyze within <think></think>. If you judge that existing knowledge is insufficient to answer accurately, proceed to search.
2. If you need to find legal basis, use <search>Query</search> to call the search. Results are returned in <information></information>.
3. You may search multiple times until information is sufficient.
4. Finally, provide a concise answer within <answer></answer>.

Question:"""
\end{promptbox}
\end{figure*}

\FloatBarrier

\begin{figure*}[htbp]
\begin{promptbox}{Summarization Prompt}
summarization_prompt = """You are a legal information extraction expert. Please extract legal information points relevant to the user's question from the following webpage content.

User Question:
{original_question}

Requirements:
1. Only extract legal information directly related to the question (articles, regulations, cases, interpretations, etc.).
2. Include:
   - Relevant legal articles and regulation names.
   - Specific legal provisions and requirements.
   - Scope of application, conditions, and exceptions.
   - Relevant legal consequences or liabilities.
3. Do not analyze the question, do not give conclusions, and do not add your own reasoning.
4. Remain objective and neutral; directly extract legal facts.
5. Remove all webpage URL links.
6. Summarize concisely in 6-10 sentences.
7. If the content is unrelated to the question, directly state "The webpage content is unrelated to the question".

Webpage Content:
{content}
"""
\end{promptbox}
\end{figure*}

\FloatBarrier